\documentclass[a4paper,conference]{IEEEtran}
\usepackage{amsmath,graphicx,bm,amssymb,bbding}
\usepackage{booktabs}
\usepackage{multirow}
\usepackage{url}
\usepackage{xcolor}
\usepackage{soul}
\soulregister\cite7
\soulregister\citep7
\soulregister\citet7
\soulregister\ref7
\soulregister\pageref7
\usepackage[normalem]{ulem}
\useunder{\uline}{\ul}{}
\IEEEoverridecommandlockouts
\ifCLASSINFOpdf
\else
\fi
\begin{document}

\title{Exploiting Spatial-temporal Correlations for Video Anomaly Detection}

\author{\IEEEauthorblockN{Mengyang Zhao}
\IEEEauthorblockA{Academy for Eng. \& Tech.\\
Fudan University\\
Shanghai, China\\
myzhao20@fudan.edu.cn}
\and
\IEEEauthorblockN{Yang Liu}
\IEEEauthorblockA{Academy for Eng. \& Tech.\\
Fudan University\\
Shanghai, China\\
yang\_liu20@fudan.edu.cn}
\and
\IEEEauthorblockN{Jing Liu}
\IEEEauthorblockA{Academy for Eng. \& Tech.\\
Fudan University\\
Shanghai, China\\
jingliu19@fudan.edu.cn}
\and
\IEEEauthorblockN{Xinhua Zeng*\thanks{\hrule \vspace{3pt}*Corresponding author. This work was supported by the Shanghai Key Research Lab of NSAI}}
\IEEEauthorblockA{Academy for Eng. \& Tech.\\
Fudan University\\
Shanghai, China\\
zengxh@fudan.edu.cn}
}

\maketitle

\begin{abstract}
  Video anomaly detection (VAD) remains a challenging task in the pattern recognition community due to the ambiguity and diversity of abnormal events. Existing deep learning-based VAD methods usually leverage proxy tasks to learn the normal patterns and discriminate the instances that deviate from such patterns as abnormal. However, most of them do not take full advantage of spatial-temporal correlations among video frames, which is critical for understanding normal patterns. In this paper, we address unsupervised VAD by learning the evolution regularity of appearance and motion in the long and short-term and exploit the spatial-temporal correlations among consecutive frames in normal videos more adequately. Specifically, we proposed to utilize the spatiotemporal long short-term memory (ST-LSTM) to extract and memorize spatial appearances and temporal variations in a unified memory cell.
  In addition, inspired by the generative adversarial network, we introduce a discriminator to perform adversarial learning with the ST-LSTM to enhance the learning capability. Experimental results on standard benchmarks demonstrate the effectiveness of spatial-temporal correlations for unsupervised VAD. Our method achieves competitive performance compared to the state-of-the-art methods with AUCs of 96.7\%, 87.8\%, and 73.1\% on the UCSD Ped2, CUHK Avenue, and ShanghaiTech, respectively.

  \textit{Index terms—}Video anomaly detection, adversarial learning, spatial-temporal correlations, spatial-temporal consistency
\end{abstract}

\IEEEpeerreviewmaketitle

\section{Introduction}
Video anomaly detection (VAD) has been intensively studied because of its potential to be used in intelligent surveillance systems \cite{ramachandra2020survey}. The goal of VAD is to identify unfamiliar or unexpected events in a specific scenario, such as traffic accidents and violence \cite{liu2022collaborative}. Since that abnormal events are diverse and difficult to acquire, there are many methods \cite{liu2018future,cai2021appearance,chang2020clustering,liu2022appearance} trained on only normal samples. They learn the normality in an unsupervised manner and consider the instances that deviate from the normal pattern as abnormal ones.

In recent years, many VAD methods \cite{liu2018future,chang2022video,nguyen2019anomaly} focusing on spatial and temporal information have been proposed. They tend to learn normal patterns utilizing spatial-temporal representations and achieve superior performance. They generally take a sequence of video frames as part of the input. However, they have difficulty learning spatial appearances and temporal variations in the long and short-term, limited by their backbone network. Most of them take optical flow as temporal information, which only tends to capture the variation regularity between two adjacent frames instead of the complete motion evolution \cite{nguyen2019anomaly}. Moreover, they usually consider spatial and temporal information separately while ignoring the correspondence between them, i.e., the spatial-temporal consistency. On the one hand, the spatial-temporal correlations among frames can help the model understand the motion evolution of objects. For example, it may be abnormal if the position of a car changes too much in some consecutive frames. On the other hand, spatial-temporal consistency considers the correspondence between appearance and motion signals, e.g., a person walking in front of a store is normal, but it could be abnormal to walk on a green belt. The two points mentioned above are crucial for VAD. However, most of the existing methods do not concentrate on them.

In this paper, we explore the evolution regularity of appearance and motion in consecutive frames utilizing spatiotemporal long short-term memory (ST-LSTM) to help the model execute the frame prediction, which serves as the proxy task for VAD. The ST-LSTM is improved based on the Convolutional LSTM (ConvLSTM) \cite{xingjian2015convolutional}, which models appearance and motion representations in the long and short-term so that our model exploits the spatial-temporal correlations in context frames more adequately. In addition, our method accounts for spatial-temporal consistency because spatial and temporal representations are captured in the same memory cell. Therefore, the proposed spatial-temporal correlations network (STC-Net) learns normal patterns more comprehensively, decreasing the prediction error of normal instances, hence being more capable of distinguishing abnormal from normal. We also propose to use the bidirectional sequence as input to extract richer spatial-temporal correlations and introduce a novel strategy named error accumulation, which enlarges the prediction error for abnormal events. Additionally, adversarial training \cite{goodfellow2014generative} is employed to promote the model prediction capability. The main contributions of this paper are summarized as follows:
\begin{itemize}
  \item Our method exploits spatial-temporal correlations among frames more adequately and accounts for spatial-temporal consistency, which provides a new way of using spatial-temporal information in unsupervised VAD. 
  \item Besides, some general components have been devised for VAD, which drives abnormal instances more separated from normal patterns, and their effectiveness has been verified in the experiment. 
  \item The experiments on benchmark datasets indicate the proposed method is competitive with state-of-the-art methods. In addition, we provide the comprehensive experimental analysis including ablation studies and visualization analysis.
\end{itemize}

\section{Related work}
\subsection{Reconstruction-based Video Anomaly Detection}
 The reconstruction-based methods have been proposed \cite{hasan2016learning,luo2017revisit,DBLP:conf/icip/RavanbakhshNSMR17,gong2019memorizing} in recent years, which are trained on only normal events and consider the instances difficult to reconstruct as abnormal. For example, Hasan \textit{et al.} \cite{hasan2016learning} introduced a deep autoencoder to learn the regularity of normal video sequences. They use autoencoder to reconstruct video frames and consider the instances with high reconstruction error as abnormal. Luo \textit{et al.} \cite{luo2017revisit} proposed a temporally-coherent sparse coding to force neighboring frames reconstructed by similar reconstruction coefficients. However, sometimes deep autoencoders can reconstruct the abnormal frames in low errors as well because of the overgeneralization. To this end, Gong \textit{et al.} \cite{gong2019memorizing} devised a memory module for autoencoder, which constrains the ability to reconstruct abnormal frames by memorizing the prototypical elements of the normal frames. Although the above methods are somewhat efficient in learning normal patterns, they do not take advantage of the spatial-temporal information in context frames, which is critical for the model to understand normal patterns.

\subsection{Spatial and Temporal Representations in VAD }
Recently, many VAD methods \cite{liu2018future,chang2022video,cai2021appearance,nguyen2019anomaly,DBLP:conf/icmcs/LaiLH20,liu2022learning} exploiting spatial and temporal representations have been proposed. They usually achieve superior performance compared to the reconstruction-based methods \cite{hasan2016learning,luo2017revisit,DBLP:conf/icip/RavanbakhshNSMR17,gong2019memorizing}. For example, Liu \textit{et al.} \cite{liu2018future} proposed the prediction-based VAD method, which utilizes several consecutive frames as input and predicts a future frame. At testing time, instances with higher prediction error are considered as abnormal. Their proposed Frame-Pred model considers optical flow as motion (temporal) constraint and performs the prediction task by a fine-tuned U-Net \cite{ronneberger2015u}. In addition, many two-stream models have been proposed, which typically utilize two branches to learn spatial and temporal representations separately. Chang \textit{et al.} \cite{chang2022video} devised a two-stream model including spatial and temporal autoencoders to dissociate the spatial-temporal representations. The model executes the reconstruction task using spatial autoencoder to learn appearance representation, while the prediction task is performed using a motion autoencoder based on U-Net to learn motion representation. Cai \textit{et al.}  \cite{cai2021appearance}  attempted to explore spatial-temporal consistency and proposed a two-stream model named AMMC-Net. They designed two memory pools for the model, one with frames as input for learning appearance representation and the other with the optical flow for learning motion representation. Although the above methods have made efforts in learning spatial and temporal representations and achieve better performance, they almost have several shortcomings in exploiting spatial-temporal correlations. For example, Frame-Pred \cite{liu2018future} and Chang \textit{et al.} \cite{chang2022video} utilize U-Net \cite{ronneberger2015u} as the feature extractor, which is not adept at learning temporal variations, and they pay little attention to spatial-temporal consistency. Nguyen \textit{et al.} \cite{nguyen2019anomaly} and AMMC-Net \cite{cai2021appearance} consider optical flow as temporal information, which could not reflect the motion evolution ultimately. 

\begin{figure}[t]
  \centering
  \includegraphics[width=.45\textwidth]{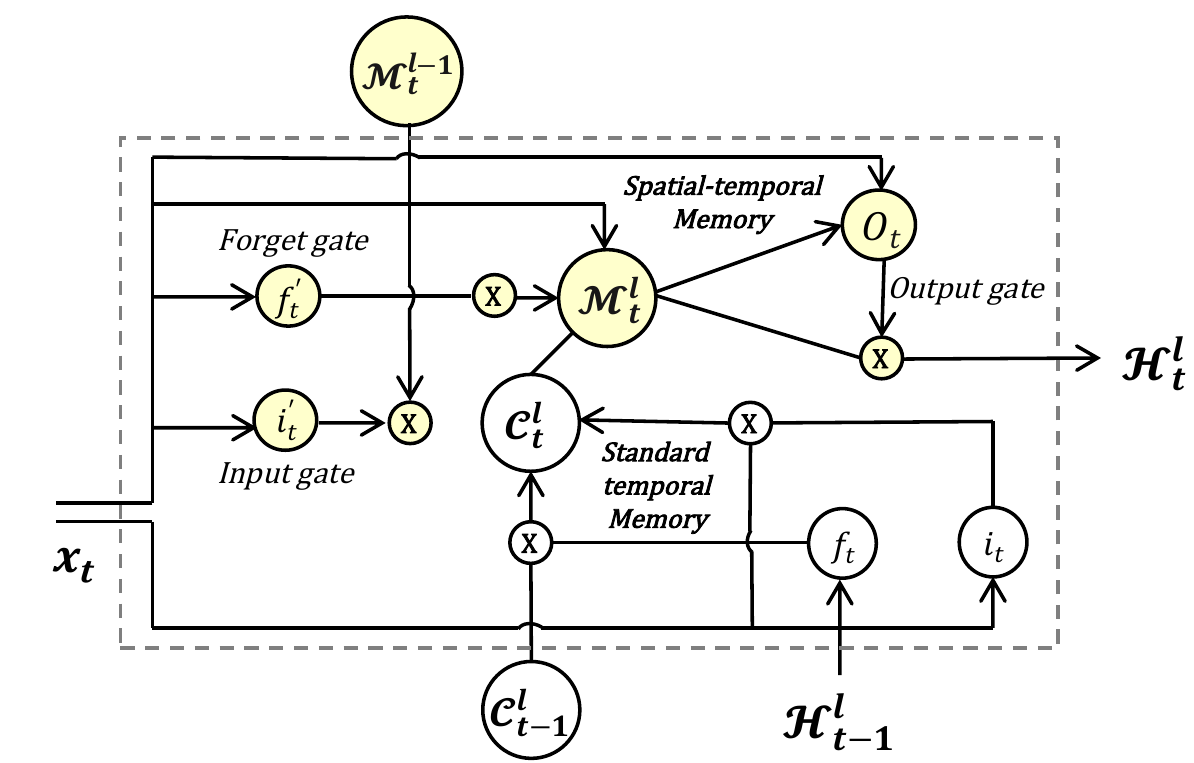}
  \caption{Details of the ST-LSTM unit. The yellow circles indicate the spatial-temporal memory, which are the differences compared with the ConvLSTM. \bm{$i_t$} and \bm{$f_t$} denote the input gate and forget gate of $\bm{\mathcal{C}}^l_t$, respectively. Similarly, $\bm{i}^\prime_t$ and $\bm{f}^\prime_t$ are the input gate and forget gate of $\bm{\mathcal{M}}^l_t$.}
  \label{fig:1}
\end{figure}
\begin{figure*}
  \centering
  \includegraphics[width=.99\textwidth]{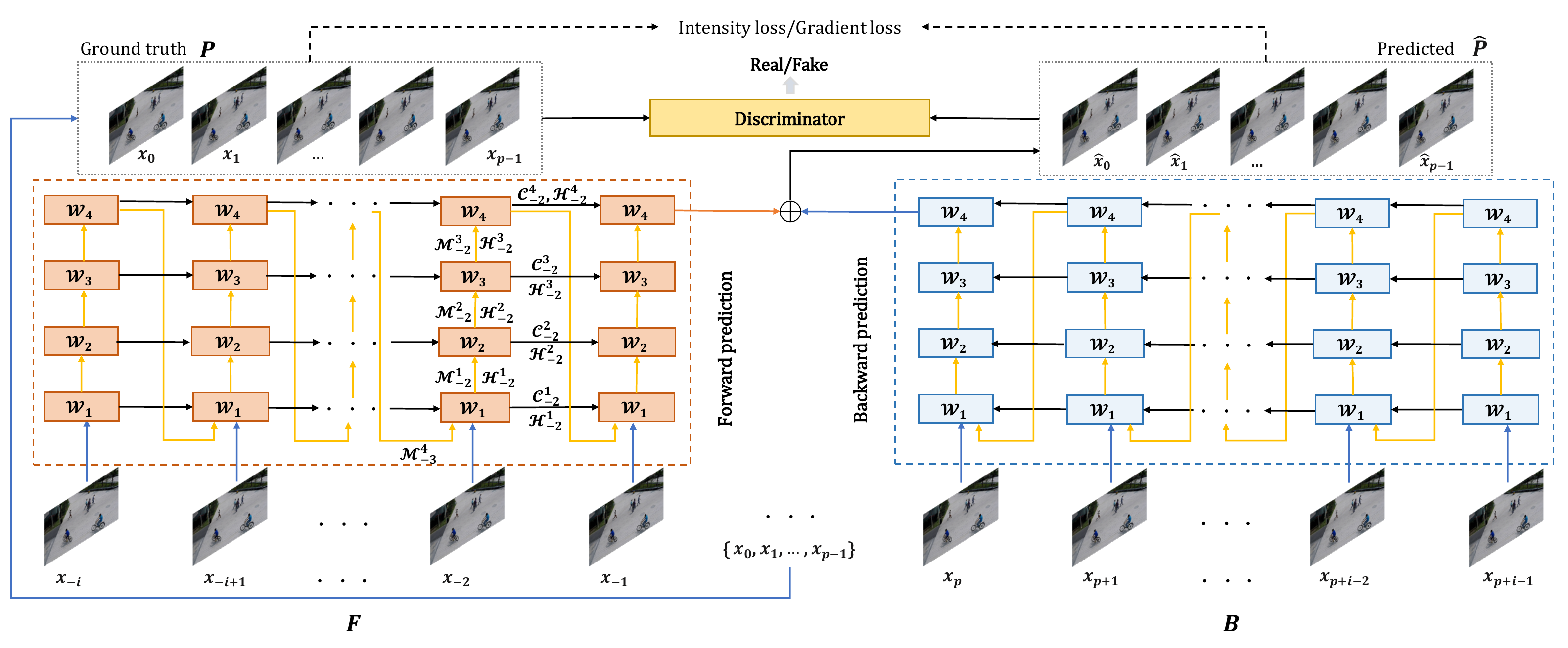}
  \caption{The overall structure of the STC-Net. The generator is a convolutional recurrent neural network consisting of four ST-LSTM units. $\bm{\mathcal{W}}_i$ denotes the convolutional parameters. $\bm{\mathcal{C}}^l_t$ and $\bm{\mathcal{M}}^l_t$ are the standard temporal memory and the spatial-temporal memory, respectively. $\bm{\mathcal{H}}^l_t$ is the hidden state. The yellow arrows denote the transition path of spatial-temporal memory $\bm{\mathcal{M}}^l_t$, while the black arrows denote the standard temporal memory $\bm{\mathcal{C}}^l_t$ flow.}
  \label{fig:2}
\end{figure*}
\section{Method}
The prediction-based VAD methods \cite{liu2018future,ye2019anopcn,park2020learning} consider the instances with lower prediction errors as normal while considering unpredictable events as abnormal. Therefore, the prediction capability is significant for the performance. Inspired by video prediction research \cite{wang2017predrnn,wu2021greedy}, we propose to utilize ST-LSTM as the generator, helping our model learn normal patterns more adequately. The ST-LSTM overcomes some disadvantages of ConvLSTM \cite{xingjian2015convolutional}, which has obvious advantages in learning spatial-temporal correlations and considering spatial-temporal consistency. Inspired by the generative adversarial network (GAN) \cite{goodfellow2014generative,liu2021generative}, we devised adversarial training to force the predicted frames closer to the ground truth. The bidirectional prediction strategy is designed so that the model learns richer spatial-temporal correlations from the context frames. We also introduce a novel strategy named error accumulation, which could enlarge the prediction error in abnormal events. In conclusion, our approach improves the prediction ability for normal instances by exploiting spatial-temporal correlations adequately and drives abnormal instances to deviate more from normal patterns by the devised components.

\subsection{The Details of ST-LSTM}
The ConvLSTM network \cite{xingjian2015convolutional} has been already extended to video prediction and VAD tasks \cite{wang2017predrnn,DBLP:conf/icassp/LeeKR18,feng2021convolutional} with remarkable performance. However, in this network, features are updated only in the vertical and horizontal directions which results in the features memorized by the top memory cell being ignored by the bottom layer at the next time step \cite{wang2017predrnn}. To overcome this disadvantage, a novel ST-LSTM is devised for the spatial-temporal sequence forecasting model. As shown in Figure 2, the ST-LSTM is devised with the zigzag direction update strategy based on the ConvLSTM. The spatial features can be delivered from the top layer to the bottom layer of the next time step. Therefore, ST-LSTM can model spatial and temporal representations in a unified memory cell. It overcomes the layer-independent memory mechanism in ConvLSTM and achieves better performance in video sequence prediction. The details of ST-LSTM unit are shown in Figure~\ref{fig:1}. Its equations are as follows: 
\begin{flalign}
  \bm{\mathcal{C}}_{t}^{l} & =\bm{f}_{t} \odot \bm{\mathcal{C}}_{t-1}^{l}+\bm{i}_{t} \odot \tanh (\bm{\mathcal{W}}_{x g} * \bm{x}_{t} \nonumber \\ &\quad +\bm{\mathcal{W}}_{h g} *\bm{ \mathcal{H}}_{t-1}^{l}+\bm{b}_{g})\\
 \bm{ \mathcal{M}}_{t}^{l} & = \bm{f}_{t}^{\prime} \odot \bm{\mathcal{M}}_{t}^{l-1}+\bm{i}_{t}^{\prime} \odot \tanh (\bm{\mathcal{W}}_{\bm{x} g}^{\prime} *\bm{ x}_{t} \nonumber \\ & \quad +\bm{\mathcal{W}}_{m g} * \bm{\mathcal{M}}_{t}^{l-1}+\bm{b}_{g}^{\prime})\\
  \bm{o}_{t} & =\sigma(\bm{\mathcal{W}}_{x o} * \bm{x}_{t}+\bm{\mathcal{W}}_{h o} * \bm{\mathcal{H}}_{t-1}^{l}+\bm{\mathcal{W}}_{c o} * \bm{\mathcal{C}}_{t}^{l} \nonumber \\ & \quad +\bm{\mathcal{W}}_{m o} * \bm{\mathcal{M}}_{t}^{l}+\bm{b}_{o})\\
  \bm{\mathcal{H}}_{t}^{l} & = \bm{o}_{t} \odot \tanh \left(\bm{\mathcal{W}}_{1 \times 1} *\left[\bm{\mathcal{C}}_{t}^{l}, \bm{\mathcal{M}}_{t}^{l}\right]\right)
\end{flalign}

where * and $\mathcal\odot$ denote the convolution operator and the Hadamard product, respectively, and $\mathcal\sigma$ is the sigmoid activation function.



As shown in Figure~\ref{fig:2}, similar to ConvLSTM, the temporal cell $\bm{\mathcal{C}}_{t}^{l}$ is updated horizontally from the time step $t-1$ to the next time step $t$ within all layers. $\bm{\mathcal{M}}_{t}^{l}$ is the unified memory cell mentioned above, which models spatial and temporal features. For the bottom of ST-LSTM, the spatial-temporal memory conveyed from the top layer at the time step $t-1$ to the bottom layer at the next time step $t$, $\bm{\mathcal{M}}_{t}^{0}=\bm{\mathcal{M}}_{t-1}^{L}$. For the other layers, it is transmitted from the $l-1$ layer to the $l$ layer vertically. Finally, the hidden states $\bm{\mathcal{H}}_{t}^{l}$ is fused by the temporal cell and the spatial-temporal cell so that it can effectively memorize spatial appearances and temporal variations in the long and short-term.

\subsection{Overall Structure}
The overall structure of our model is shown in Figure~\ref{fig:2}. We devised the generator based on ST-LSTM, inspired by \cite{wang2017predrnn}, which generates the predicted frames by considering neighboring frames. The discriminator is devised to enforce the predicted frames closer to the ground truth. Mathematically, we defined four sequences \bm{$F$}, \bm{$P$}, \bm{$\hat{P}$} and \bm{$B$}, where \bm{$F}=\left\{\bm{x}_{-i}, \bm{x}_{-i+1}, \bm{x}_{-i+2}, \ldots, \bm{x}_{-1}\right\}$, $ \bm{P}=\left\{\bm{x}_{0},\bm{ x}_{1}, \bm{x}_{2}, \ldots,\bm{ x}_{p-1}\right\}$, $ \bm{\hat{P}}=\left\{\bm{\hat{x}}_{0}, \bm{\hat{x}}_{1}, \bm{\hat{x}}_{2}, \ldots,\bm{\hat{x}}_{p-1}\right\}$ and $ \bm{B}=\left\{\bm{x}_{p}, \bm{x}_{p+1}, \bm{x}_{p+2}, \ldots, \bm{x}_{p+i-1}\right\}$. The sequence \bm{$P$ } and \bm{$\hat{P}$} are the ground truth and predicted sequence, respectively, including $p$ inter-frames. The sequence \bm{$F$} and \bm{ $B$ } are used for forward and backward prediction, respectively, containing $i$ frames as inputs of the generator. Most proposed methods \cite{liu2018future,cai2021appearance} tend to generate the predicted sequence only considering the sequence $\bm{F}$ in front of the ground truth. Such predictions seem to have some drawbacks because the associations between \bm{$P$} and \bm{$B$} are ignored. Therefore, we devised the bidirectional prediction, which helps the model to learn richer spatial-temporal correlations from context frames. The sequences \bm{$F$} is input to the forward prediction network while the sequences \bm{$B$} are input to the backward prediction network. As described in the previous subsection, the yellow arrows in figure \ref{fig:2} indicate the spatial-temporal memory $\bm{\mathcal{M}}_{t}^{l}$ transition path, while the black arrows indicate the transition path of standard temporal memory $\bm{\mathcal{C}}_{t}^{l}$. Thus, the final hidden state memorizes the evolution regularity appearance and motion in the long and short-term, considering the spatial-temporal correlations adequately. We concatenate the hidden states from the forward and backward prediction networks together. Then the final output is obtained by applying a $1 \times 1$ convolutional layer for dimensionality reduction.

\subsection{Adversarial Training and Constraints}
To further improve the performance, adversarial training and constraints are devised to minimize the difference between the predicted frames and ground truth. GAN has been employed numerous times for video or image generation with remarkable success \cite{liu2018future,liu2021generative,chen2021nm}. We leverage the Least Square GAN \cite{mao2017least} and treat the prediction network based on ST-LSTM as the generator $\mathcal{G}$. 

Similar to \cite{DBLP:conf/cvpr/IsolaZZE17}, we utilize a patch discriminator $\mathcal{D}$ . The purpose of training the discriminator is to distinguish between predicted sequence \bm{$\hat{P}$} and ground truth \bm{$P$}, classifying \bm{$\hat{P}$} into class 0 (False) and \bm{$P$} into class 1 (True). Therefore, the loss function of the discriminator is imposed:
\begin{equation}
  \begin{aligned}
  L_{a d v}^{\mathcal{D}}(\bm{\hat{P}},\bm{ P})&=\sum_{i, j} \frac{1}{2} L_{M S E}\left(\mathcal{D}(\bm{P})_{i, j}, 1\right) \\
  &\quad +\sum_{i, j} \frac{1}{2} L_{MSE}\left(\mathcal{D}(\bm{\hat{P}})_{i, j}, 0\right)
  \end{aligned}
  \end{equation}
where $i$ and $j$ denote the spatial patches indexes. And the mean squared error (MSE) function $L_{MSE}$ is defined as follows:
\begin{equation}
  L_{M S E}(\bm{\hat{y}}, \bm{y})=(\bm{\hat{y}}-\bm{y})^{2}
\end{equation}
The generator expects to generate more realistic \bm{$\hat{P}$}, to defeat the discriminator, causing the discriminator to classify \bm{$\hat{P}$} into class 1 (True). Therefore, the loss function of the generator is designed as follows:
\begin{equation}
  L_{a d v}^{\mathcal{G}}(\bm{\hat{P}})=\sum_{i, j} \frac{1}{2} L_{M S E}\left(\mathcal{D}(\bm{\hat{P})}_{i, j}, 1\right)
\end{equation}

In addition to GAN, we also utilize the constraints on intensity and gradient inspired by Frame-Pred \cite{liu2018future}. The constraint on intensity ensures the similarity of the pixels in \bm{$\hat{P}$} and \bm{$P$}, in RGB space. Mathematically, the intensity loss can be written as:
\begin{equation}
  L_{int}(\bm{\hat{P}},\bm{ P})= L_{M S E}(\bm{\hat{P}}, \bm{P})
\end{equation}
The constraint on gradient can sharpen every frame in \bm{$\hat{P}$}, the gradient loss is defined as follows:
\begin{equation}
    L_{g d}(\bm{\hat{P}},\bm{P})=\sum_{k=0}^{p-1}L_{g d}(\bm{\hat{x}}_{k},\bm{x}_{k})
\end{equation}
\begin{equation}
\begin{aligned}
    L_{g d}(\bm{\hat{x}},\bm{ x})&=\sum_{i, j}\left\|\left|\bm{\hat{x}}_{i, j}-\bm{\hat{x}}_{i-1, j}\right|-\left|\bm{x}_{i, j}-\bm{x}_{i-1, j}\right|\right\|_{1} \\ & \quad
    +\left\|\left|\bm{\hat{x}}_{i, j}-\bm{\hat{x}}_{i, j-1}\right|-\left|\bm{x}_{i, j}-\bm{x}_{i, j-1}\right|\right\|_{1}
    \end{aligned}
\end{equation} 
where $\bm{x}_{k}$, $\bm{\hat{x}}_{k}$ denote a frame in $\bm{P}$ and the corresponding predicted frame of it and $i$, $j$ denote the position index of pixels.

Finally, inspired by \cite{wang2021predrnn}, the decouple loss is utilized in our model. It motivates \bm{$\mathcal{C}$} and \bm{$\mathcal{M}$} to learn separate features by extending the distance between them in latent space, avoiding the two memory states to be intertwined \cite{wang2021predrnn}. The decouple loss is defined as follows:
\begin{equation}
  \Delta \bm{\mathcal{C}}_{t}^{l} =\bm{W}_{c} *\left(\bm{i}_{t} \odot \bm{g}_{t}\right)
\end{equation}
\begin{equation}
  \Delta \bm{\mathcal{M}}_{t}^{l} =\bm{W}_{c} *\left(\bm{i}_{t}^{\prime} \odot\bm{ g}_{t}^{\prime}\right)
\end{equation}
\begin{equation}
  L_{\text {decouple }} =\sum_{t, l} \frac{\left|\left\langle\Delta\bm{ \mathcal{C}}_{t}^{l}, \Delta \bm{\mathcal{M}}_{t}^{l}\right\rangle\right|}{\left\|\Delta \bm{\mathcal{C}}_{t}^{l}\right\|_{2}\left\|\Delta \bm{\mathcal{M}}_{t}^{l}\right\|}_{2}
\end{equation}
where $\bm{W} _ {c}$ is $1 \times 1$ convolutions in ST-LSTM units, and $\langle\cdot, \cdot\rangle$ denotes dot product.
\subsection{Total Loss}
The objective function takes into account all the constraints and the adversarial training. Finally, the total loss can be written as
\begin{equation}
  \begin{aligned}
    L_{\bm{\mathcal{G}}} &=\lambda_{i n t} L_{i n t}(\bm{\hat{P}},\bm{ P})+\lambda_{g d} L_{g d}(\bm{\hat{P}}, \bm{P}) \\
    & \quad +\lambda_{a d v} L_{a d v}^{\mathcal{G}}(\bm{\hat{P}})+\lambda_{d e c} L_{\text {decouple }}
    \end{aligned}
\end{equation}
When we train discriminator, the follow function is used:
\begin{equation}
  L_{\mathcal{D}}=L_{a d v}^{\mathcal{D}}(\bm{\hat{P}}, \bm{P}) 
\end{equation}

\subsection{The Error Accumulation Strategy }
The prediction-based VAD methods tend to consider instances with higher prediction error as abnormal. Therefore, we propose the error accumulation strategy, which can enlarge the prediction error in abnormal events. If the input contains abnormal frames, the prediction error will accumulate gradually. For example, only considering forward prediction for simplicity, when we use \bm{$F$} containing abnormal frame to generate $\bm{\hat{x}}_{0}$, there will be a high prediction error between $\bm{x}_{0}$ and $\bm{\hat{x}}_{0}$. Then, when we try to generate $\bm{\hat{x}}_{1}$, the errors will be accumulated, and we will get a higher prediction error because $\bm{\hat{x}}_{0}$ will be required as part of the input for the generator. At the testing time, all frames in \bm{$P$} are invisible to the generator, and we use the already generated frames as part of the input to generate subsequent frames. Therefore, instead of using the current prediction error to calculate the regular score, we use the accumulated error of the subsequent predicted frames. 

The regular score is usually calculated based on the difference between predicted frames and real frames. Mean squared error is a common way to measure the difference. However, we evaluate the quality of predicted frames utilizing the Peak Signal to Noise Ratio (PSNR), which has proven to be a more effective measure \cite{park2020learning}. The PSNR is shown as follows:
\begin{equation}
  \operatorname{P}(\bm{x}, \bm{\hat{x}})=10 \log _{10} \frac{\left[\max _{\bm{\hat{x}}}\right]^{2}}{\frac{1}{N} \sum_{i=0}^{N}\left(\bm{x}_{i}-\bm{\hat{x}}_{i}\right)^{2}}
\end{equation}
Following previous work \cite{liu2018future,park2020learning}, The PSNR of each test frame is normalized to the range $[0,1]$ and the following formula calculates the regular score of each frame:
\begin{equation}
  S(t)=\frac{\operatorname{P}\left(\bm{x}_{t'}, \bm{\hat{x}}_{t'}\right)-\min _{t'} \operatorname{P}\left(\bm{x}_{t'}, \bm{\hat{x}}_{t'}\right)}{\max _{t'} \operatorname{P}\left(\bm{x}_{t'}, \bm{\hat{x}}_{t'}\right)-\min _{t'} \operatorname{P}\left(\bm{x}_{t'}, \bm{\hat{x}}_{t'}\right)}
\end{equation}
Notably, we use the accumulated predicted error of the predicted frame $\bm{\hat{x}}_{t'}$ to calculate the regular score of $\bm{x}_{t}$, where $\bm{\hat{x}}_{t'} \in \bm{\hat{P}}$.

\section{Experiments}
In this section, we provide extensive experiment analysis, including comparison experiments with state-of-the-art methods, visualization analysis, and ablation studies. At first, in comparison experiments, we demonstrate the significance of exploring spatial-temporal correlations for the VAD methods. Secondly, the visualization analysis validates our model has superior prediction capability by learning the evolution regularity of appearance and motion. Finally, the ablation studies demonstrate the effectiveness of adversarial training, bidirectional prediction, and error accumulation strategy.
\subsection{Datasets and Evaluation Metric}
\noindent \textbf{Datasets. } We conduct experiments on three unsupervised VAD benchmark datasets, including UCSD Ped2 \cite{mahadevan2010anomaly}, CUHK Avenue \cite{lu2013abnormal} and ShanghaiTech \cite{luo2017revisit}, which are described below.
\begin{itemize}
   \item \textbf{UCSD Ped2} \cite{mahadevan2010anomaly} contains 16 training and 12 testing  videos. The dataset is captured on pedestrian walkways, defining events such as bicycling and skateboarding as anomalies.
  \item \textbf{CUHK Avenue} \cite{lu2013abnormal} dataset includes 37 videos consisting of about 30,000 frames in total. There are 21 testing videos including abnormal events such as fast running and throwing.
  \item \textbf{ShanghaiTech} \cite{luo2017revisit} dataset consists of 330 training videos and 107 test videos from 13 different scenes. It contains many complex anomalies, such as fighting, robbing, etc.
\end{itemize}

\noindent \textbf{Evaluation metrics.} 
To quantitatively evaluate our method, following the previous work \cite{gong2019memorizing}, the average area under the curve (AUC) is utilized as the evaluation metrics.

\subsection{Implementation Details}
In experiments, all frames are resized to $256 \times 256$ and the intensity of pixels is normalized to [0, 1]. For the CUHK Avenue \cite{lu2013abnormal} and ShanghaiTech \cite{luo2017revisit} dataset, the learning rate of the generator and discriminator are set to $3\times 10^{-4}$ and $3\times 10^{-5}$, respectively. For the UCSD Ped2 \cite{mahadevan2010anomaly} dataset, they start from $1\times 10^{-4}$ and $1\times 10^{-5}$, separately. $\lambda_{int}, \lambda_{g d}, \lambda_{a d v}$ and $\lambda_{\text {dec }}$ are set to $1.0$, $1.0$, $0.05$ and $1.0$, respectively. In addition, we used the Adam \cite{kingma2014adam} to optimize the networks and a batch size of 4. We set $i$ and $p$ to 8 and 5, respectively, which means the generator uses eight frames before \bm{$P$} and eight frames after \bm{$P$ } to generate \bm{ $\hat{P}$} containing five frames. Finally, inspired by \cite{wang2021predrnn}, the reverse scheduled sampling strategy used in training phase, which utilizes part of the predicted frames from normal events as input to generate subsequent frames. This strategy enables the model to memorize long-term evaluation patterns from the context frames.

In testing, we calculate the normal score frame by frame, calculating the score of the first frame in \bm{$P$} each time only. In addition, $t'$ is set to $t+2$ , which means that we use the accumulated error PSNR $\left(\bm{x}_{t+2}, \bm{\hat{x}}_{t+2}\right)$ to calculate the regular score of $\bm{x}_{t}$.

\subsection{Comparison with State-of-the-art Methods}
\begin{table}[]
  \caption{Results of quantitative frame-level AUC (\%) comparison. Bold Numbers indicate the best performance, while underlie ones indicate the second best.}
  \label{tab:1}
  \resizebox{.49\textwidth}{!}{
  \begin{tabular}{@{}llccc@{}}
  \toprule
  \multicolumn{2}{c}{\textbf{Method}}   & \multicolumn{1}{l}{\textbf{UCSD Ped2}} & \multicolumn{1}{l}{\textbf{CUHK Avenue}} & \multicolumn{1}{l}{\textbf{ShanghaiTech}} \\ \midrule
  \multirow{3}{*}{\rotatebox[]{90}{\textbf{Trad.}}}  & MPPCA \cite{kim2009observe}      & 69.3         & N/A   & N/A    \\
         & MPCC+SFA \cite{kim2009observe}     & 64.3         & N/A   & N/A    \\
         & MDT \cite{mahadevan2010anomaly}        & 82.9         & N/A   & N/A    \\ \midrule
  \multirow{4}{*}{\rotatebox[]{90}{\textbf{Recon.}}} & Conv-AE \cite {hasan2016learning}         & 85.0         & 80.0  & 60.9   \\
         & TSC \cite{luo2017revisit} & 91.0         & 80.6  & 67.9   \\
         & StackRNN \cite{luo2017revisit}        & 92.2         & 81.7  & 68.0   \\
         & MemAE \cite{gong2019memorizing}  & 94.1         & 83.3  & 71.2   \\ \midrule
  \multirow{6}{*}{\rotatebox[]{90}{\textbf{Spat.-Temp.}}}    & ConvLSTM-AE \cite{luo2017remembering}     & 88.1         & 77.0  & N/A    \\
         & Frame-Pred   \cite{liu2018future}     & 95.4         & 84.9  & 72.8   \\
         & Nguyen \textit{et al.} \cite{nguyen2019anomaly} & 96.2         & 86.9  & N/A    \\
         & Chang \textit{et al.} \cite{chang2022video} & \textbf{96.7}    & {\ul 87.1}     & 73.7   \\
         & AMMC-Net \cite{cai2021appearance}         & {\ul 96.6}   & 86.6  & \textbf{73.7}   \\ \cmidrule(l){2-5} 
         & STC-Net (Ours)   & \textbf{96.7}    & \textbf{87.8}  & {\ul 73.1}      \\ \bottomrule
  \end{tabular}}
  \end{table}
We report the frame-level AUC of the proposed STC-Net on three benchmark VAD datasets mentioned above \cite{mahadevan2010anomaly}, \cite{lu2013abnormal}, \cite{luo2017revisit}. The methods involved in the comparison include traditional models (Trad.) \cite{kim2009observe,mahadevan2010anomaly}, reconstruction-based models (Recon.) \cite{luo2017revisit,hasan2016learning,gong2019memorizing} and models using spatial-temporal representations (Spat.-Temp.) \cite{liu2018future,nguyen2019anomaly,chang2022video,cai2021appearance,luo2017remembering}. The results are presented in Table~\ref{tab:1}. Compared with the traditional and reconstruction methods in Table~\ref{tab:1}, STC-Net achieves a significant performance advantage due to exploiting spatial-temporal correlations. Additionally, the remaining methods in  Table~\ref{tab:1} have made lots of efforts in learning spatial and temporal representations. However, all of them are difficult to learn the evolution regularity of appearance and motion in the long and short-term. Therefore, STC-Net is more competitive compared with them.

\subsection{Visualization Analysis}
\begin{figure}[t]
  \centering
  \includegraphics[width=.49\textwidth]{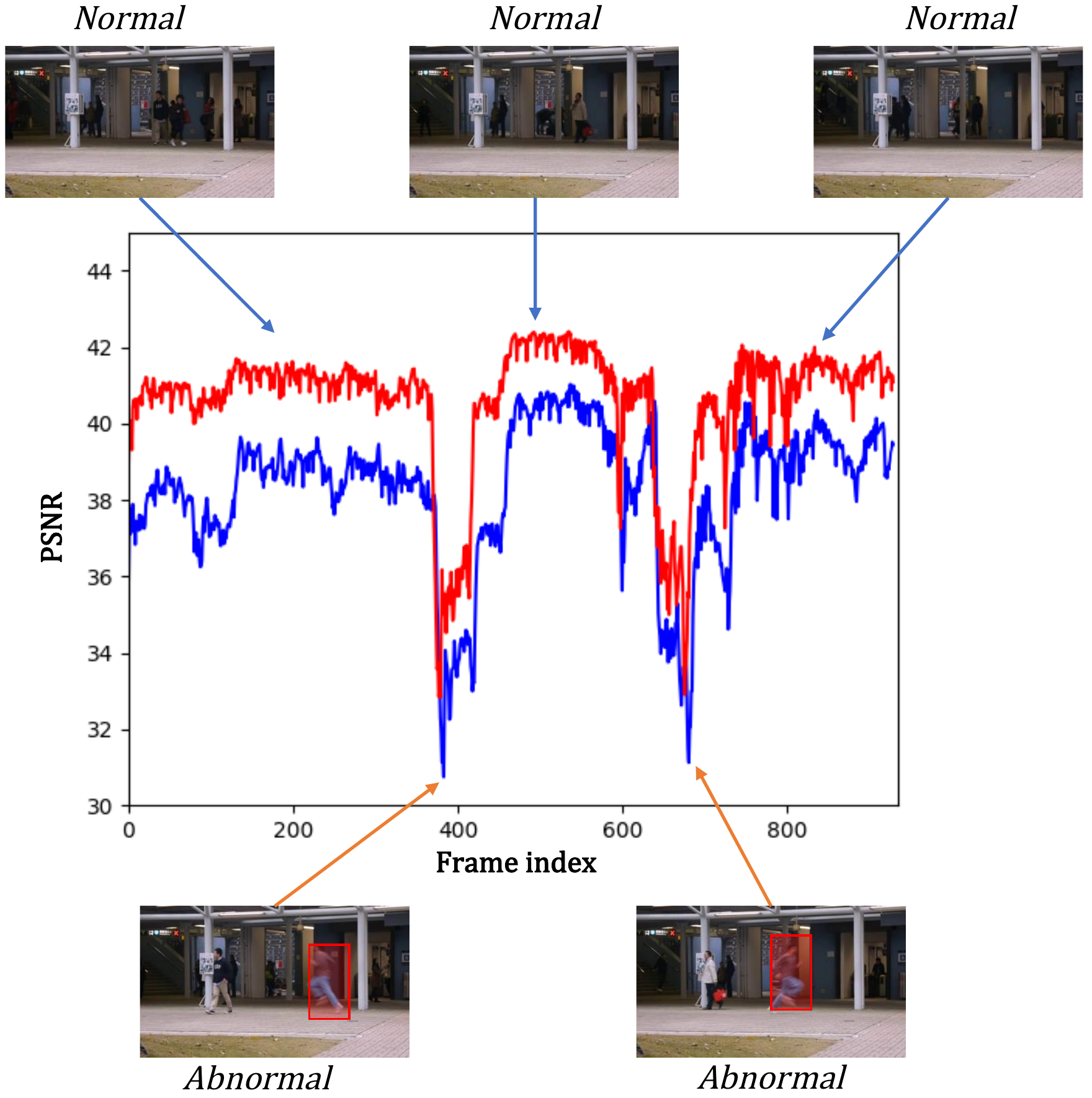}
  \caption{Part of the temporal PSNR between predicted frames and ground truth on the Avenue dataset. The red curve represents the proposed STC-Net and the blue curve represents the Frame-Pred \cite{liu2018future} .}
  \label{fig:3}
\end{figure}
\begin{figure}[b]
  \centering
  \includegraphics[width=.49\textwidth]{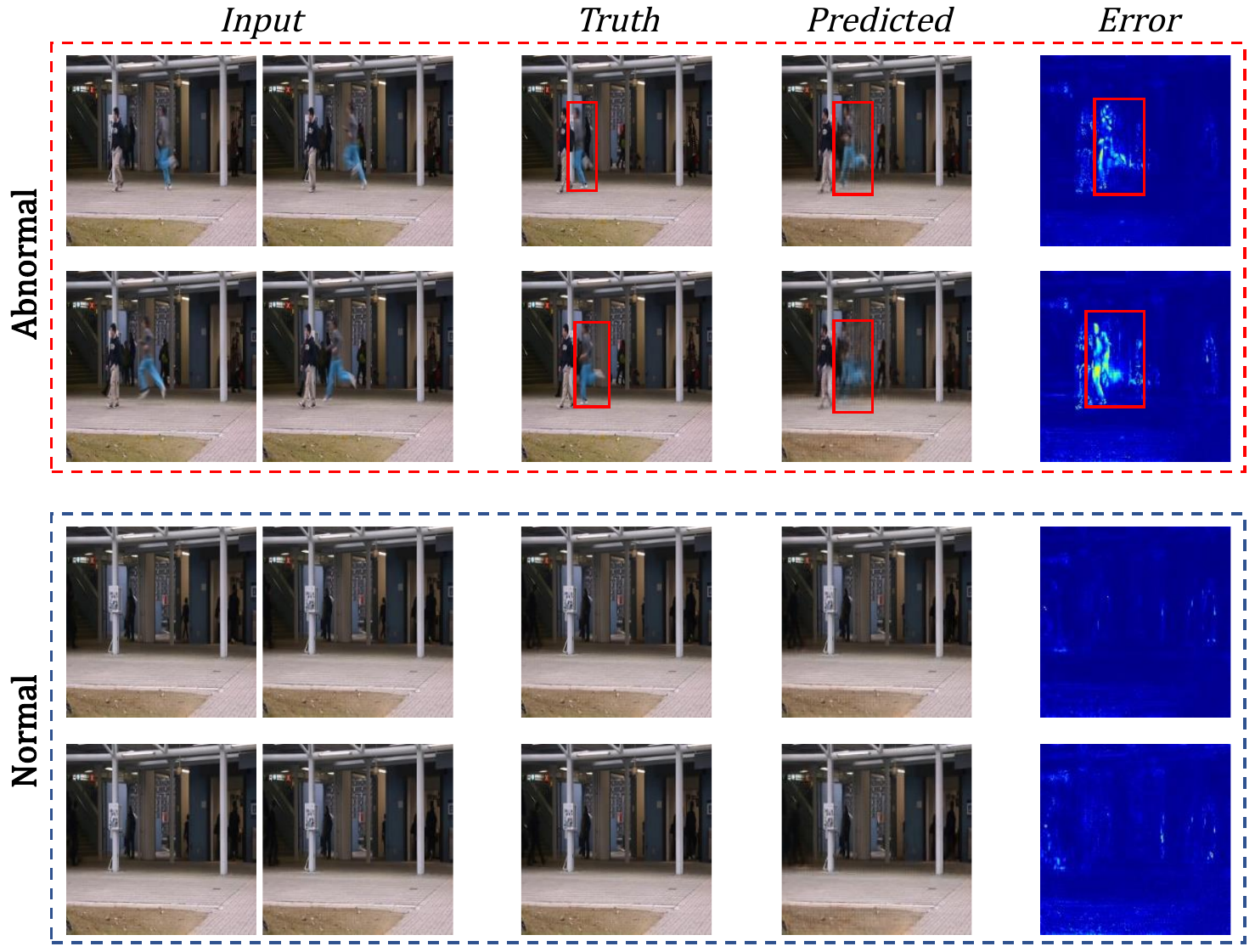}
  \caption{Visualization of predicted frames and prediction errors. For simplicity, only part of frames are shown, the four input frames $\left\{\bm{x}_{-8},\bm{ x}_{-6}, \bm{x}_{-4},  \bm{x}_{-2}\right\}$ are shown, from left to right and top to bottom. The ground truth is frames $\left\{\bm{x}_{0},\bm{ x}_{2}\right\}$ from top to bottom, and its right side is the corresponding predicted frames 
  $\left\{\bm{\hat{x}}_{0},\bm{\hat{x}}_{2}\right\}$. The rightmost column represents the difference between the predicted frames and ground truth, meaning the prediction error.}
  \label{fig:4}
\end{figure}
The enhancement of the spatial-temporal correlations on the prediction ability is verified, which is reflected in the predicted frames of our model are closer to the ground truth compared to Frame-Pred \cite{liu2018future}. In addition, we visualize the prediction errors under normal and abnormal events, respectively. The visualization results provide spatial and temporal localization of abnormal events and further validate the effectiveness of the devised components.

The model considers unpredictable instances as abnormal. Thus the lower PSNR in Figure~\ref{fig:3} implies a higher prediction error, corresponding to the period of an abnormal event. The predicted frames closer to the ground truth have higher PSNR values, corresponding to the normal instances. As shown in Figure~\ref{fig:3}, the PSNR curve of our model in normal events is higher and smoother compared to Frame-Pred \cite{liu2018future}. While in abnormal events, they are almost equivalent. In addition, we compute the gap between average PSNR of normal frames and that of abnormal frames as statistical metric, denoted as $ \Delta P$. The STC-Net achieves $ \Delta P$ = 2.9, while the Frame-Pred  \cite{liu2018future} achieves $ \Delta P$ = 2.6 on UCSD ped2 \cite{mahadevan2010anomaly}. This further validates that our model is superior in distinguishing between normal and abnormal events.

According to Figure~\ref{fig:4}, we can observe that the prediction error is very obvious in abnormal events, corresponding to the upper part of Figure~\ref{fig:4}. However, the predicted frames are remarkably close to the ground truth in the normal events, corresponding to the lower part of Figure~\ref{fig:4}. In addition, we can noticeably observe that the prediction error in the second row is considerably higher than in the first row. However, the prediction error in the fourth row is almost indistinguishable from the third row. This is because the second row shows the difference between \bm{${x}_{2}$} and \bm{ $\hat{x}_{2}$}, which is the accumulated error. This further validates the effectiveness of the error accumulation strategy, which can enlarge the prediction error under abnormal events with little impact on normal events.

\subsection{Ablation Studies}
\begin{table}[]
  \centering
  \caption{Quantitative comparison for variants of our model. We measure the frame-level AUC on the CUHK avenue dataset. Bold number indicates the best performance.}
  \label{tab:2}
  \resizebox{.49\textwidth}{!}{
  \begin{tabular}{@{}ccccc@{}}
  \toprule
  \textbf{Model} & \textbf{GAN} & \textbf{Bidirectional prediction} & \textbf{Error accumulation} & \textbf{AUC(\%)} \\ \midrule
  1 &  \XSolidBrush         &  \XSolidBrush    &  \XSolidBrush  & 85.6    \\
  2 & \Checkmark         &  \XSolidBrush    &  \XSolidBrush  & 86.6    \\
  3 & \Checkmark         & \Checkmark    &  \XSolidBrush  & 87.1    \\ \midrule
  STC-Net        & \Checkmark         & \Checkmark    & \Checkmark  & \textbf{87.8}    \\ \bottomrule
  \end{tabular}}
\end{table}

To demonstrate the effectiveness of the devised components, the ablation analysis is devised as shown in Table~\ref{tab:2}. We report the frame-level AUC for the variants of STC-Net on CUHK Avenue \cite{lu2013abnormal} dataset. Model 1 is the baseline model, which utilizes only forward prediction and does not include adversarial training. Compared to Model 1, Model 2 adds adversarial training and enhances AUC performance by 1\%. Model 3 utilizes the bidirectional prediction strategy based on Model 2, bringing a 0.5\% improvement. STC-Net is the final model, utilizing the error accumulation strategy based on Model 3, enhancing the AUC performance by 0.7\%. The results of ablation studies show that each of the devised components is significant for the performance of our model.
\section{Conclusion}
In this paper, we propose a spatial-temporal correlation network to address unsupervised video anomaly detection by exploring spatial-temporal correlations among frames utilizing ST-LSTM. The proposed STC-Net explores the evolution regularity of appearance and motion in the long and short-term and considers spatial-temporal consistency. Experimental results on benchmark datasets demonstrate that our STC-Net is competitive with state-of-the-art methods. The visualization results further validate the effectiveness of spatial-temporal correlations for VAD. Additionally,  the effectiveness of the devised components, such as the error accumulation strategy, are demonstrated by ablation studies. In future work, we will investigate applying spatial-temporal correlations to other models, such as two-stream models, to achieve superior performance.

\bibliographystyle{IEEEtran}

\bibliography{IEEEabrv,mylib}
\end{document}